\pdfoutput=1
\documentclass[11pt]{article}

\usepackage[final]{acl}

\usepackage{times}
\usepackage{latexsym}
\usepackage[T1]{fontenc}
\usepackage[utf8]{inputenc}
\usepackage{microtype}
\usepackage{inconsolata}

\usepackage{graphicx}
\usepackage{booktabs}
\usepackage{tcolorbox}
\usepackage{todonotes}

\title{Big City Bias: Evaluating the Impact of Metropolitan Size on Computational Job Market Abilities of Language Models}

\author{Charlie Campanella \\
  Indeed \\
  \texttt{ccampanella@indeed.com} \\\And
  Rob van der Goot \\
  IT University of Copenhagen \\
  \texttt{robv@itu.dk} \\}

\begin{document}
\maketitle
\begin{abstract}
Large language models (LLMs) have emerged as a useful technology for job matching, for both candidates and employers. Job matching is often based on a particular geographic location, such as a city or region. However, LLMs have known biases, commonly derived from their training data. In this work, we aim to quantify the metropolitan size bias encoded within large language models, evaluating zero-shot salary, employer presence, and commute duration predictions in 384 of the United States’ metropolitan regions. Across all benchmarks, we observe negative correlations between the metropolitan size and the performance of the LLMS, indicating that smaller regions are indeed underrepresented. More concretely, the smallest 10 metropolitan regions show upwards of 300\% worse benchmark performance than the largest 10.\footnote{https://github.com/charlie-campanella/big-city-bias}
\end{abstract}

\section{Introduction}
\label{sec:intro}
Recent large language models (LLMs) are primarily trained on internet-derived corpora~\cite{brown2020language,touvron2023llama}. These underlying datasets are prone to linguistic and geographic bias. For example, the training data of Common Crawl, used to train OpenAI’s GPT-3~\cite{brown2020language} and Meta AI’s Llama~\cite{touvron2023llama}, is composed of 46\% English language documents. Such lexical imbalances contribute to an anglophone bias in various tasks, exemplified by GPT-3.5’s "English-first" approach when translating Irish/Gaeilge~\cite{chiarainfilling} and its inability to pass Indonesian primary school exams~\cite{koto-etal-2023-large} while simultaneously passing U.S. college-entry exams~\cite{openai2023gpt4}. Evidently, under-representation in training corpora can adversely affect language model performance across various tasks and contexts.

A significant population disparity exists among metropolitan regions in the United States, with the largest, New York–Newark–Jersey City, NY-NJ, having over 300 times more residents than the smallest, Eagle Pass, TX. These population disparities seem to correlate with the amount of associated information available online. For instance, querying “New York, NY” on Wikipedia yields 114,067 results while a search for “Eagle Pass, TX” only returns 1,516 results. The same queries on Google result in 1.7 billion and 8.2 million results, respectively. This training data disparity prompts us to consider whether language models exhibit comparable biases in performance, excelling in tasks associated with larger metropolitan regions.

In this work, we quantify the "big city bias" by evaluating salary, employer presence, and commute duration predictions in the 384 metropolitan statistical areas (MSAs) defined by the United States Census Bureau. Across all benchmarks, analysis indicates a  correlation between MSA population and predictive accuracy, indicating superior language model efficacy in the context of larger cities. Given the concentration of technical talent in the United States' 10 most populous metropolitan areas \footnote{https://www.bls.gov/oes/current/oes151252.htm}, these findings might inspire AI practitioners to look beyond so-called "tech hubs" when applying LLM-generated synthetic data to geographically diverse job matching applications.

We contribute:
\begin{itemize}
    \item A dataset with population, employer presence, and commute duration for 384 metropolitan areas in the USA.
    \item Outputs of 5 recent language models predicting employer presence, commute duration, and salary.
    \item An analysis of the results of the language models for this data and task, showing that larger metropolitan areas are more closely associated with accurate predictive outcomes.
\end{itemize}

\section{Methodology}
\subsection{Data and tasks}
\label{sec:setup}

Job matching systems are designed to find the best-fitting role for a given candidate. For example, Indeed.com, the world's most popular job site \footnote{\url{https://www.indeed.com/about}}, considers a role to be a "bad fit" if there is a salary or geographic mismatch, \footnote{\url{https://engineering.indeedblog.com/blog/2019/09/jobs-filter/}} among other criterion. Our geographically-focused evaluation prompts language models to make \textit{salary}, \textit{employer presence}, and \textit{commute duration} predictions for each metropolitan region to evaluate their performance at common matching tasks.

Prediction biases can reduce the effectiveness of LLM-powered job matching systems. For example, a system which fails to predict salaries within a given region may match job seekers with financially incompatible roles. Inaccurate commute duration forecasts can lead job seekers to consider positions that are logistically unfeasible. Furthermore, not discerning whether an employer's regional presence is a major facility or a branch office can mislead job seekers about career progression prospects.

\begin{figure}
    \includegraphics[width=\columnwidth]{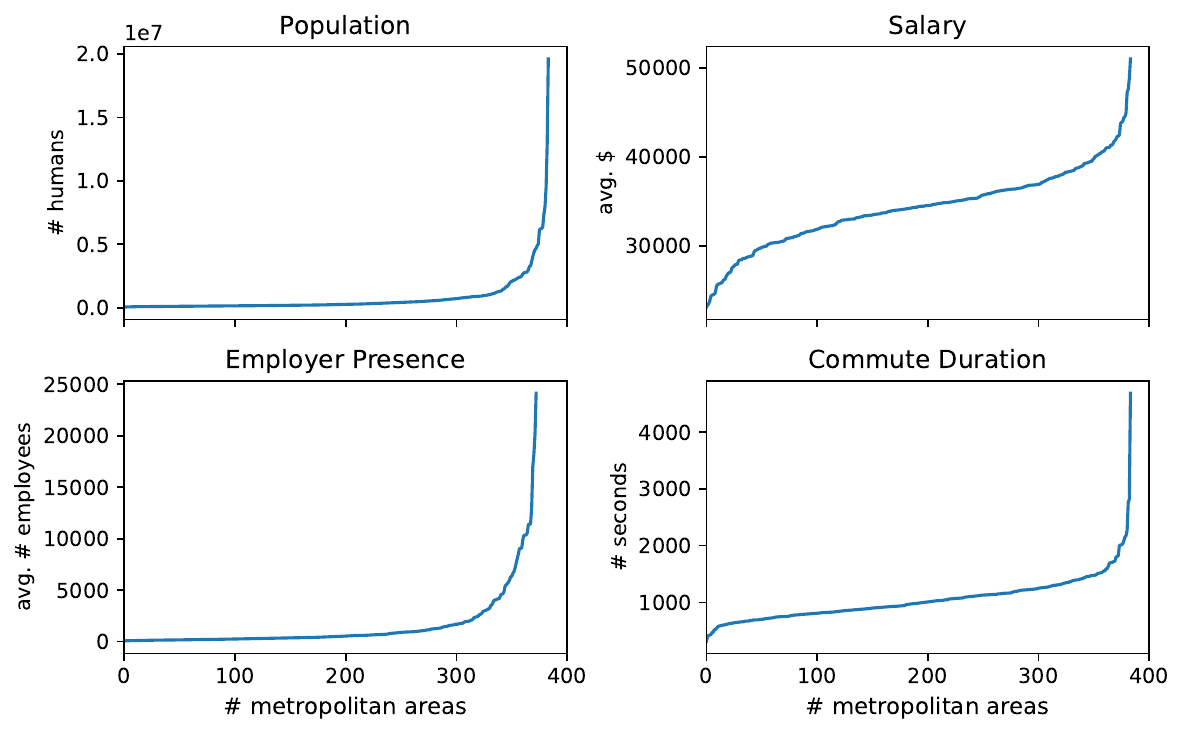}
    \caption{For each variable the cumulative number of metropolitan areas which match a certain threshold (y-axis). Note that the target variables (all except population) are an average over multiple instances within the region, and are thus not directly comparable (i.e.\ they could be over different sets of jobs, employers, or commutes).}
    \label{fig:distributions}    
\end{figure}

As mentioned in Section~\ref{sec:intro}, this analysis is conducted on metropolitan areas as defined by the United States Census Bureau. For each task, we leverage data collected from a reputable source, with distributions for each target variable plotted in Figure~\ref{fig:distributions}. The data for each task is described in detail below.

\paragraph{Salary Task Data:} Sourced from a confidential and proprietary dataset by Indeed.com which includes pre-tax, base annual salaries for registered nurses, software engineers, and warehouse workers in each MSA.

\paragraph{Employer Presence Task Data:}
Based on a publicly accessible dataset published by People Data Labs \footnote{https://www.peopledatalabs.com/top-employers-dataset} which presents the top 10 employers and their respective employment counts across 384 U.S. metropolitan regions. Our modified version of this dataset incorporates additional "state\_code" and "metro\_state\_code" columns, aligning with MSA naming conventions used by the U.S. Census Bureau for standardized referencing.

\paragraph{Commute Duration Task Data:} Derived from a custom script which uses the Google Maps API to generate 5 random commute origins, destinations, and durations within 20km of the geographic center for each MSA. Note that each duration estimate assumes driving as the  mode of transport.

\begin{figure}
    \centering
    \includegraphics[width=.7\columnwidth]{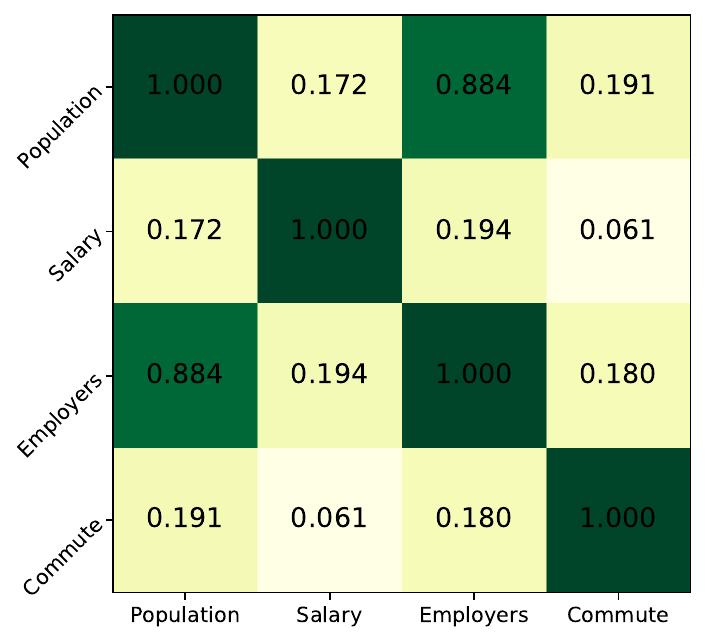}
    \caption{Pearson correlations between MSA population and each target variable. Note that the bottom-left and top-right are mirrored as Pearson correlations are not directional.}
    \label{fig:eda}
\end{figure}

We plot the correlation of the population size and each of the target variables against each other in Figure~\ref{fig:eda}. From this, we can clearly see that the average employer presence and MSA population have a very strong correlation; this makes intuitive sense, as there is greater possibility in larger cities to have larger employers. Other combinations of variables, especially salary and commute duration, show less significant correlation, although the slopes in Figure~\ref{fig:distributions} have similar shapes.

\subsection{Models}
\label{sec:models}
Our selection of language models was optimized for diversity in architectures, training methods, and parameter size while minimizing computational costs. Based on these criterion, we selected \textit{mistral-instruct-7b-v0.1}~\cite{jiang2023mistral}, \textit{llama-2-chat-7b}, \textit{llama-2-chat-70b}~\cite{touvron2023llama2}, \textit{gpt-3.5-turbo}~\cite{brown2020language}, and \textit{gpt-4}~\cite{openai2023gpt4}.

We employ these models using prompting, designing task-specific prompts which contain: the question, relevant entities for the task, and instructions on how to structure the answer, so that the numerical response can be automatically extracted. All prompts are shown below:

\begin{tcolorbox}[title=Prompt 2.1: Salary Predictions]
\begin{flushleft}
    What is the average annual salary for a \{JOB\_TITLE\} in \{METRO\_AREA\}? Return one estimate with no other text, do not return a range. Salary:
\end{flushleft}
\end{tcolorbox}

\begin{tcolorbox}[title=Prompt 2.2: Employer Presence Predictions]
\begin{flushleft}
How many people does \{EMPLOYER\_NAME\} employ in \{METRO\_AREA\}? Provide an estimate even if this information is not publicly available. Return one estimate with no other text, do not return a range. Number employed:
\end{flushleft}
\end{tcolorbox}

\begin{tcolorbox}[title=Prompt 2.3: Commute Duration Predictions]
\begin{flushleft}
What is the average driving commute time from \{ORIGIN\} to \{DESTINATION\} in \{METRO\_AREA\}? Provide an estimate even though you are not a GPS. Return an estimate in minutes with no other text. Commute Estimate:
\end{flushleft}
\end{tcolorbox}

\subsection{Metrics}
After obtaining a response from the language models (Section~\ref{sec:models}), we first extract numerical values through the application of a regular expression (\verb|[^0-9\.]+$|). We compare these numeric values to the gold labels (Section~\ref{sec:setup}), and then compute the percentage error for each task. For each task category in a metropolitan region, we take into account 3-5 model outputs and derive an average percentage error. This smooths out individual output variances, ensuring a more reliable evaluation.

Next, we look at the \textit{Pearson correlation}~\cite{pearson1901liii} between the log of the population size and the average prediction error. This will tell us whether larger metropolitan areas consistently have more accurate predictions or vice versa. 
We then employ the \textit{median error}, as a quantitative metric of the performance of the models. Since this is taken over the values of the variables we predict, and they vary across tasks (Section~\ref{sec:setup}), we use our third metric: \textit{coefficient of determination (r$^2$)}. This metric explains the proportion of the variation in the error that is predictable from the metropolitan size, and normally ranges from 0-1.

\section{Results}

\begin{table}
\setlength{\tabcolsep}{3pt}
\begin{tabular}{l l r r r}
\toprule
LLM & Size & Correl. & Median & r$^2$ \\
& & & \multicolumn{1}{l}{Err.\%} \\
\midrule 
  \multicolumn{1}{l}{Salary} \\
\midrule
mistral-instruct & 7b & -.0890$^*$ & 21.95 & .0079 \\
llama-2-chat & 7b & -.3553$^*$ & 22.92 & .1262 \\
llama-2-chat & 70b & -.2765$^*$ & 19.47 & .0765 \\
gpt-3.5-turbo & 20b & -.3398$^*$ & 19.49 & .1155 \\ 
gpt-4 & ? & -.4113$^*$ & 17.75 & .1692 \\
\midrule
\multicolumn{1}{l}{Employer Presence}  \\
\midrule
mistral-instruct & 7b & -.2125$^*$ & 182.60 & .0452 \\
llama-2-chat & 7b & -.3307$^*$ & 201.56 & .1094 \\
llama-2-chat & 70b & -.2731$^*$ & 149.36 & .0746 \\
gpt-3.5-turbo & 20b & -.3004$^*$ & 161.33 & .0902 \\
gpt-4 & ? & -.3179$^*$ & 143.45 & .1010 \\
\midrule
\multicolumn{1}{l}{Commute Duration}  \\
\midrule
mistral-instruct & 7b & -.1422$^*$ & 140.22 & .0202 \\
llama-2-chat & 7b & -.1141$^*$ & 141.12 & .0130  \\
llama-2-chat & 70b & -.1915$^*$ & 126.86 & .0367 \\
gpt-3.5-turbo & 20b & -.0083$^*$ & 71.11 & .0001 \\
gpt-4 & ? & -.1872$^*$ & 51.72 & .0351 \\
\bottomrule
\end{tabular}
\caption{Pearson correlations between the log of the population and the average prediction error, median error, and coefficient of determination (r$^2$). $^*$ p $<$ 0.0001.}
\label{tab:results}
\end{table}

\begin{figure*}
\centering
\includegraphics[width=.95\textwidth]{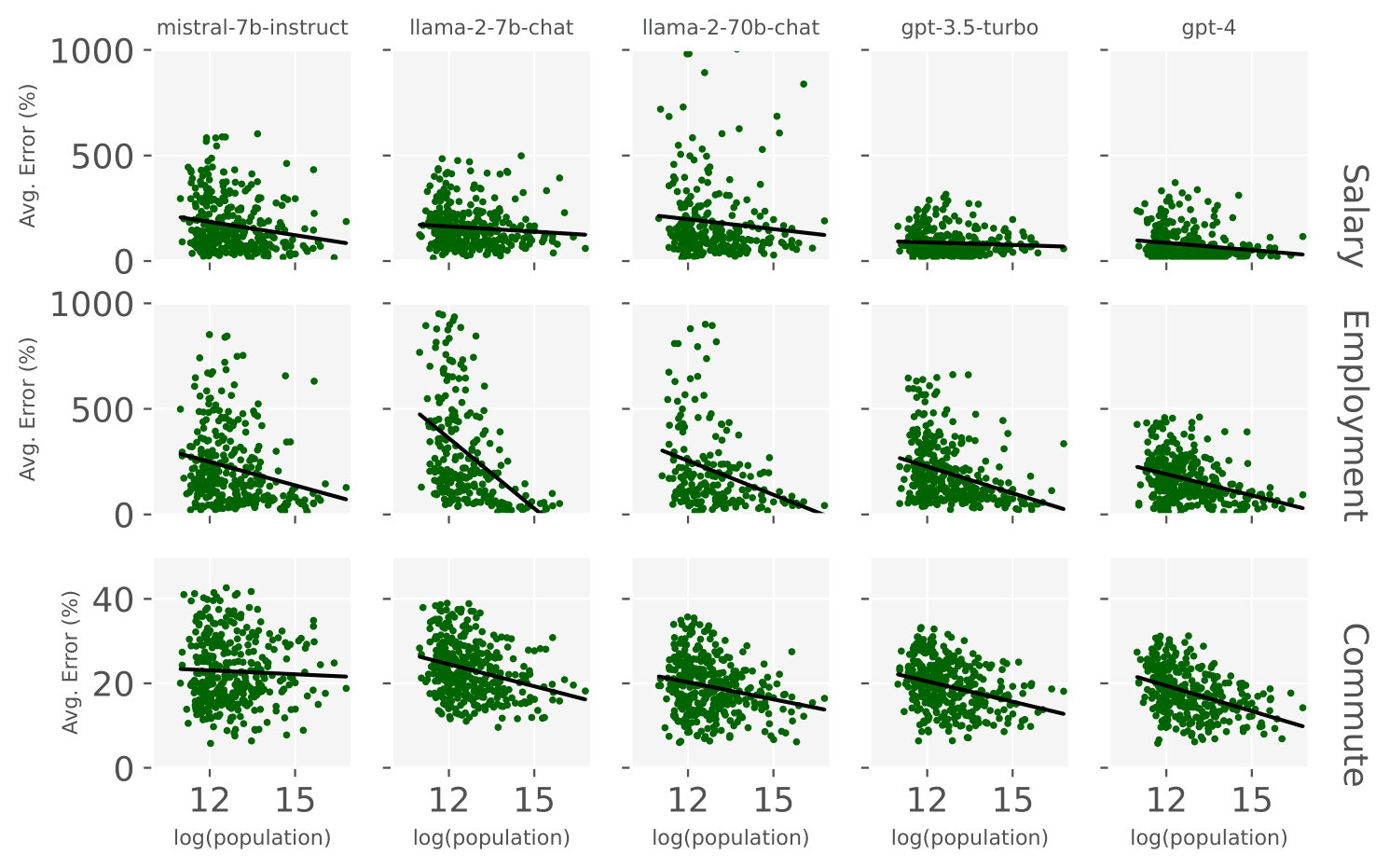}
\caption{The average error plotted against the log of the population for each task (shown on the right) and each language model (top).}
\label{fig:allResults}
\end{figure*}

\begin{table}
\setlength{\tabcolsep}{3pt}
\begin{tabular}{l l r r r}
\toprule
LLM & Size & Top10 & Bottom10 & Diff.\% \\
& & \multicolumn{1}{l}{Err.\%} & \multicolumn{1}{l}{Err. \%} \\
\midrule 
  \multicolumn{3}{l}{Salary} \\
\midrule
mistral-instruct & 7b & 23.4 & 25.2 & 7.6\\
llama-2-chat & 7b& 17.4 & 27.2 & 56.2\\
llama-2-chat & 70b& 14.2 & 20.1 & 41.9\\
gpt-3.5-turbo & 20b& 15.0 & 23.3 &  55.0\\
gpt-4 & ?& 13.2 & 23.4 & 77.0\\
\midrule 
  \multicolumn{3}{l}{Employer Presence} \\
\midrule
mistral-instruct& 7b & 240.7 & 330.7 & 37.4\\
llama-2-chat & 7b& 46.4 & 467.6 & 906.9\\
llama-2-chat & 70b& 169.5 & 631.9 &  272.8\\
gpt-3.5-turbo & 20b& 100.1 & 323.0 & 222.9\\
gpt-4 & ?&68.6 & 244.2 & 256.0\\
\midrule 
  \multicolumn{3}{l}{Commute Duration} \\
\midrule
mistral-instruct& 7b & 215.0 & 431.7 &  100.8\\
llama-2-chat& 7b & 155.7 & 268.9 & 72.7\\
llama-2-chat& 70b & 169.5 & 631.9 & 272.8\\
gpt-3.5-turbo& 20b & 83.4 & 188.5 & 125.9\\
gpt-4 & ? & 56.5 & 232.8 & 312.4\\
\bottomrule
\end{tabular}
\caption{Performance of language models on the 10 largest and 10 smallest metropolitan areas. The ``Diff.\%'' column indicates the normalized percentage increase in error when comparing the bottom 10 to the top 10. i.e.\ a diff/\% of 10 means that the error is 10\% higher for the bottom 10 as compared to the top 10.}
\label{tab:top10}
\end{table}

Experimental results are displayed in Table~\ref{tab:results}. Nearly all experiments demonstrate negative and significant Pearson coefficients, indicating that LLMs tend to achieve better performance in tasks related to larger cities. Commute duration tasks generally show less stable outcomes, with lower correlations and a broader range of median errors, whereas salary prediction tasks consistently present higher correlations and smaller median errors. Median errors are highest for employer presence predictions, which possibly relates to its distribution being the most skewed (Figure~\ref{fig:distributions}). 

\section{Analysis}
\subsection{Visualizations}
Rendering scatterplots of our predictions (Figure~\ref{fig:allResults}), we observe the same trend as in Table~\ref{tab:results}: all correlations are negative to some extent.  In these visualizations, we observe that the LLaMA series of language models have the most outliers, with the 70b parameter variant having the most trouble with larger populations (especially for salary). This suggests that larger model size does not necessarily lead to more consistent predictions.

\subsection{Top 10 vs Bottom 10}
In Table~\ref{tab:top10}, we compare the performance of the language models on the largest and smallest 10 metropolitan areas. It is evident that performance is better for tasks associated with larger regions. While target variable error magnitudes vary greatly, even the approx. 50\% average difference among salary tasks is quite noticeable. Performance disparities among commute duration and employer presence are much larger, with up to 9x worse employer presence performance in the bottom 10 metropolitan areas by \textsc{llama-2-chat-7b}.

\section{Related Work}
Research using the open-source PopQA dataset shows that LLMs exhibit worse performance at tasks associated with "popular" entities. ~\cite{mallen-etal-2023-trust} Additionally, a separate study reveals a correlation between a nation's GDP-per-capita and performance on related LLM tasks. ~\cite{10386329} While both studies emphasize disparities in LLM task performance based on popularity and/or geography, neither assesses HR-related task performance across metropolitan areas.

\section{Conclusion}
Large language models, while powerful, show sub\-optimal performance in predicting salaries, commute duration, and employer presence in specific regions, with this trend worsening in smaller areas. While LLMs seem unsuitable at generating such job matching data, practitioners should remain vigilant in mitigating geographic bias as these models undergo further development and improvement.

\section*{Limitations}
Our study's primary limitation lies in the inherent nature of language models as probabilistic systems, which leads to inconsistent outputs. For example, generating \textsc{llama-2-chat-7b} employment presence predictions only yielded 259 valid, non-outlier data points, less than other models. Additionally, our research was predominantly U.S.-centric, limiting its applicability to other geographical contexts. Another limitation is our focus on absolute error metrics, which fail to indicate whether the model systematically overestimates or underestimates certain variables, like salaries. Addressing these issues in future research could improve accuracy and applicability.

\bibliography{papers}



\end{document}